\documentclass[journal,twoside,web]{ieeecolor}
\usepackage{generic}
\usepackage{cite}
\usepackage{amsmath,amssymb,amsfonts}
\usepackage{algorithmic}
\usepackage{graphicx}
\usepackage{algorithm,algorithmic}
\usepackage{hyperref}
\usepackage{textcomp}
\usepackage{booktabs}
\usepackage{multirow}
\usepackage{stfloats}
\usepackage[utf8]{inputenc}
\usepackage{caption}
\def\BibTeX{{\rm B\kern-.05em{\sc i\kern-.025em b}\kern-.08em
    T\kern-.1667em\lower.7ex\hbox{E}\kern-.125emX}}
\begin{document}
\title{LightVesselNet: An Ultra-Lightweight Sub-100K Parameter Network for Retinal Blood Vessel Segmentation}

\author{
Shadman Sobhan and Farhana Jalil%
\thanks{The authors are with the Department of Electrical \& Electronic Engineering, Bangladesh University of Engineering and Technology (BUET), Dhaka, Bangladesh.}%
\thanks{Corresponding author: shadmansobhan114@gmail.com.}
}

\maketitle

\begin{abstract}
Retinal blood vessel segmentation plays a vital role in the early detection of diabetic retinopathy and glaucoma. While recent deep learning models have achieved great segmentation accuracy, they typically require heavy computational resources, making real-world deployment on edge devices difficult. In this paper, we propose LightVesselNet, an efficient neural network designed for retinal vessel segmentation in a resource-constrained environment. Despite containing only \textbf{75K parameters}, LightVesselNet performs competitively with much larger models. The network employs a compact encoder–decoder architecture enhanced with channel and spatial attention mechanisms, a multi-scale feature aggregation module at the bottleneck, and a subpixel upsampling strategy in the decoder. A dedicated edge residual connection preserves fine vessel detail throughout decoding. Extensive experiments on five publicly available datasets — DRIVE, STARE, CHASE\_DB1, FIVES, and HRF — yield sensitivity scores of 0.8189, 0.8499, 0.8640, 0.8634, 0.8096, and Dice coefficients of 0.8070, 0.8072, 0.8181, 0.8649, and 0.7686, respectively. LightVesselNet shows improved efficiency (Performance vs Parameter or GFlops) compared to State-of-the-Art models. Cross-dataset evaluation confirms the model's generalisation capability. Overall, LightVesselNet is a strong candidate for deployment in low-resource clinical settings and mobile screening tools. The source code is available at: https://github.com/ShadmanSobhan/LightVesselNet

\end{abstract}

\begin{IEEEkeywords}
Edge deployment, Lightweight deep learning, Medical image segmentation, Retinal vessel segmentation, Sub-100K Parameter Network

\end{IEEEkeywords}

\section{Introduction}
\label{sec:Introduction}

Retinal vascular morphology is important for the early detection of diseases such as retinopathy, glaucoma, hypertensive retinopathy, and age-related macular degeneration \cite{bib1}\cite{bib2}\cite{bib3}. Besides, vascular abnormalities often manifest years before people perceive they are losing their vision. The World Health Organisation reports that diabetic retinopathy accounts for 4.8\% of global blindness \cite{bib4}\cite{bib5}\cite{bib6}. Consequently, the early identification of vascular anomalies holds significant potential for preventing irreversible vision loss. Retinal vessel segmentation isolates the vessels from surrounding regions and helps to easily identify whether any vessel structure is abnormal \cite{bib7}. 

Traditional vessel segmentation methods used hand-crafted techniques like matched filtering, processing, and vessel enhancement filters \cite{bib8}. Although these approaches demonstrated a degree of segmentation capability, they exhibited several critical limitations, including susceptibility to noise and illumination variation, and reduced efficacy in detecting thin, low-contrast vessels. \cite{bib9}\cite{bib10}. Machine learning approaches yielded moderate improvements. But they were still limited by the quality of these handcrafted representations.

In biomedical image segmentation, Deep Convolutional Neural Networks (CNNs) brought a revolution \cite{bib10}\cite{bib11}\cite{bib12}. U-Net-based encoder-decoder architectures improved segmentation performance even more \cite{bib13}. Attention mechanisms on images introduced by Vision Transformers, and multi-scale feature fusion further enhanced segmentation \cite{bib14}\cite{bib15}. However, vessel segmentation is generally one of the toughest types of segmentation, as blood vessels are much narrower and occupy less space than the background \cite{bib3}. So, manual annotation becomes very difficult, and there was disagreement even among annotators. The disagreement made large-scale annotation even more challenging. Combining all these, there is a lack of large-scale datasets in Retinal Vessel Segmentation.

Accurately capturing complex vessel patterns present in retinal fundus photography requires computational resources and deep learning models with large parameter counts.  Besides, cross-dataset studies confirm that segmentation models often fail to generalise to new fundus images without adaptation \cite{bib11}. Conversely, computationally intensive models are ill-suited for deployment in screening devices, mobile healthcare systems, and low-resource clinical environments, such as in rural areas. 

Methods like pruning, quantisation, and knowledge distillation can reduce the size of the models. However, these methods frequently introduce additional training complexity or incur non-trivial performance degradation \cite{bib16}. Moreover, such techniques compress pre-existing architectures rather than developing models architecturally optimised for retinal vessel segmentation.

In this work, we propose \textbf{LightVesselNet}, an ultra-lightweight model with an encoder-decoder-based architecture, specifically designed for retinal blood vessel segmentation. With only 75K trainable parameters, this model performs very well in segmenting retinal vessels compared to other SOTA models in terms of performance vs parameters. The model's ultra-lightweight nature makes it suitable for deployment on edge devices and in resource-constrained environments. The main contributions of this work are summarised as follows:

\begin{itemize}
    \item We propose \textbf{LightVesselNet}, an ultra-lightweight encoder--decoder network for retinal vessel segmentation containing only approximately \textbf{75K parameters} and requiring around \textbf{1.4 GFLOPs} at $512 \times 512$ resolution. The proposed model achieves a favourable accuracy--efficiency trade-off that remains highly competitive at this parameter scale.

    \item We introduce \textbf{MicroBlockSE}, a compact feature extraction block that combines depthwise-separable convolutions, squeeze-and-excitation attention, residual connections, and DropBlock regularisation to maximise representational capability while maintaining minimal computational overhead.

    \item We design an efficient \textbf{Multi-Scale Feature Aggregation (MSFA)} bottleneck module employing parallel dilated depthwise convolution branches with quarter-channel splitting and spatial attention, enabling effective capture of retinal vessels with varying calibres without significantly increasing computational complexity.

    \item Extensive experiments conducted on different datasets demonstrate that LightVesselNet achieves competitive segmentation performance compared to models containing $10\times$ to $100\times$ more parameters. Furthermore, cross-dataset evaluation confirms the strong generalisation capability of the proposed approach on unseen domains.
\end{itemize}

The remainder of this paper is organised as follows: Section \ref{sec:Related Work} discusses related work. Section~\ref{sec:Methodology} describes the proposed network architecture, implementation, and training configuration. Section~\ref{sec:Experiments} presents experimental results, ablation studies, cross-dataset evaluation, and comparison with state-of-the-art methods. Section~\ref{sec:Conclusion} concludes the paper and discusses directions for future work.


  

\section{Related Work}
\label{sec:Related Work}

\subsection{Retinal Vessel Segmentation Methods}
 
\textbf{Traditional methods.}
Early vessel segmentation relied on hand-crafted filters and classical machine learning. Chaudhuri et al. \cite{bib8} introduced matched filtering with 2-D Gaussian kernels oriented along vessel directions, while Hoover et al.\cite{bib17} proposed piecewise threshold probing of matched filter responses. Staal et al. \cite{bib18} combined ridge-based features with a $k$-NN classifier, and Fraz et al. \cite{bib19} applied an ensemble of boosted classifiers over gradient, Gabor, and morphological features. Although effective on high-contrast images, these approaches are susceptible to noise and illumination variation and exhibit reduced performance on thin, low-contrast capillaries.
 
\textbf{CNN-based encoder--decoder networks.}
The U-Net architecture \cite{bib13} established the encoder-decoder paradigm with skip connections as the dominant framework for medical image segmentation as well as retinal vessel segmentation, enabling precise localisation with limited training data. Subsequent variants improved upon it in complementary directions: U-Net++ \cite{bib27} introduced nested dense skip connections to reduce the semantic gap between encoder and decoder; EEA-UNet \cite{bib28} widened the receptive field via dilated convolutions; Wave-Net \cite{bib29} replaced standard skip connections with multi-scale feature fusion and a denoising block; and ResDO-UNet \cite{bib30} combined residual depthwise over-parameterised convolutions with multi-scale pooling attention. S-UNet \cite{bib31} incorporated a saliency mechanism into a bridge-style U-Net, while Ouyang et al. \cite{bib10} fused dynamic long-range dependencies with retinal edge-enhancement to improve thin-vessel recall.
 
\textbf{Attention and transformer-based methods.}
Channel-wise Squeeze-and-Excitation~(SE) attention \cite{bib32} and spatial gating via the Convolutional Block Attention Module (CBAM) \cite{bib33} have been widely adopted in retinal segmentation decoders. Vision Transformers \cite{bib14} and TransUNet \cite{bib15} extended segmentation with global self-attention, but their quadratic complexity and large parameter counts preclude practical deployment on edge devices. RetinaLiteNet \cite{bib20} addressed this by embedding a lightweight multi-head attention~(MHA) block within a compact CNN encoder--decoder, achieving simultaneous blood-vessel and optic-disc segmentation with only 0.066M parameters and 2.46 GFLOPs. Li et al. \cite{bib34} proposed dual-directional pooling attention with Selective Kernel units to handle vessels of varying calibre without self-attention overhead. The multi-scale attention-guided fusion network (MAGFNet)\cite{bib35} combined feature enhancement and hybrid pooling across scales, achieving strong accuracy at the cost of higher inference time.
 
\textbf{Loss functions and training.}
The vessel-to-background ratio of approximately 1:10 motivates specialised loss functions. Tversky loss generalises the Dice coefficient by independently weighting false negatives and false positives via $\alpha$ and $\beta$, enabling sensitivity–precision trade-off tuning. Focal loss \cite{bib36} downweights confidently classified background pixels, concentrating the gradient on hard vessel examples. LightVesselNet combines both in a composite Tversky--focal objective applied to the main and two auxiliary decoder outputs.
 
\subsection{Lightweight Deep Learning Architectures}
\textbf{Efficiency by design.}
Depthwise-separable convolutions factorize a standard $k{\times}k$ convolution into a depthwise spatial filter and a pointwise channel mixer, reducing parameters by roughly $k^2$-fold. Applied to retinal segmentation, M3U-CDVAE \cite{bib37} used MobileNet-V3 as a lightweight encoder backbone with a three-stage refinement pipeline, and G-Net Light \cite{bib38} adapted a slimmed GoogLeNet for vessel segmentation, both achieving competitive accuracy with far fewer parameters than standard U-Net. ColonSegNet \cite{bib39} demonstrated that architectures originally designed for colonoscopy segmentation can transfer to retinal vessels with only five million parameters.
 
\textbf{Sub-1M retinal architectures.}
Recent work has pushed parameter budgets well below 1M. LVS-Net \cite{bib21} achieved strong sensitivity on DRIVE and STARE with 0.71M parameters. LFA-Net \cite{bib22} reduced this to 0.11M through a lite-fusion attention mechanism, and LFRA-Net \cite{bib23} reached 0.17M with focal and region-aware attention. LW~U-Net with reverse attention \cite{bib24} explored attention-augmented decoding at 1.94 M parameters. Despite these advances, most sub-1M models exhibit a sensitivity–efficiency trade-off, particularly in the detection of thin vessels, and no existing compact design simultaneously incorporates multi-scale feature aggregation, subpixel upsampling, and dedicated edge preservation.

\section{Methodology}
\label{sec:Methodology}

\subsection{Model Architecture}

LightVesselNet is an ultra-lightweight encoder–decoder network built for retinal blood vessel segmentation in a resource-constrained environment. The design philosophy prioritises parameter efficiency.

\begin{figure*}[!t]
    \centering
    \includegraphics[width=\linewidth]{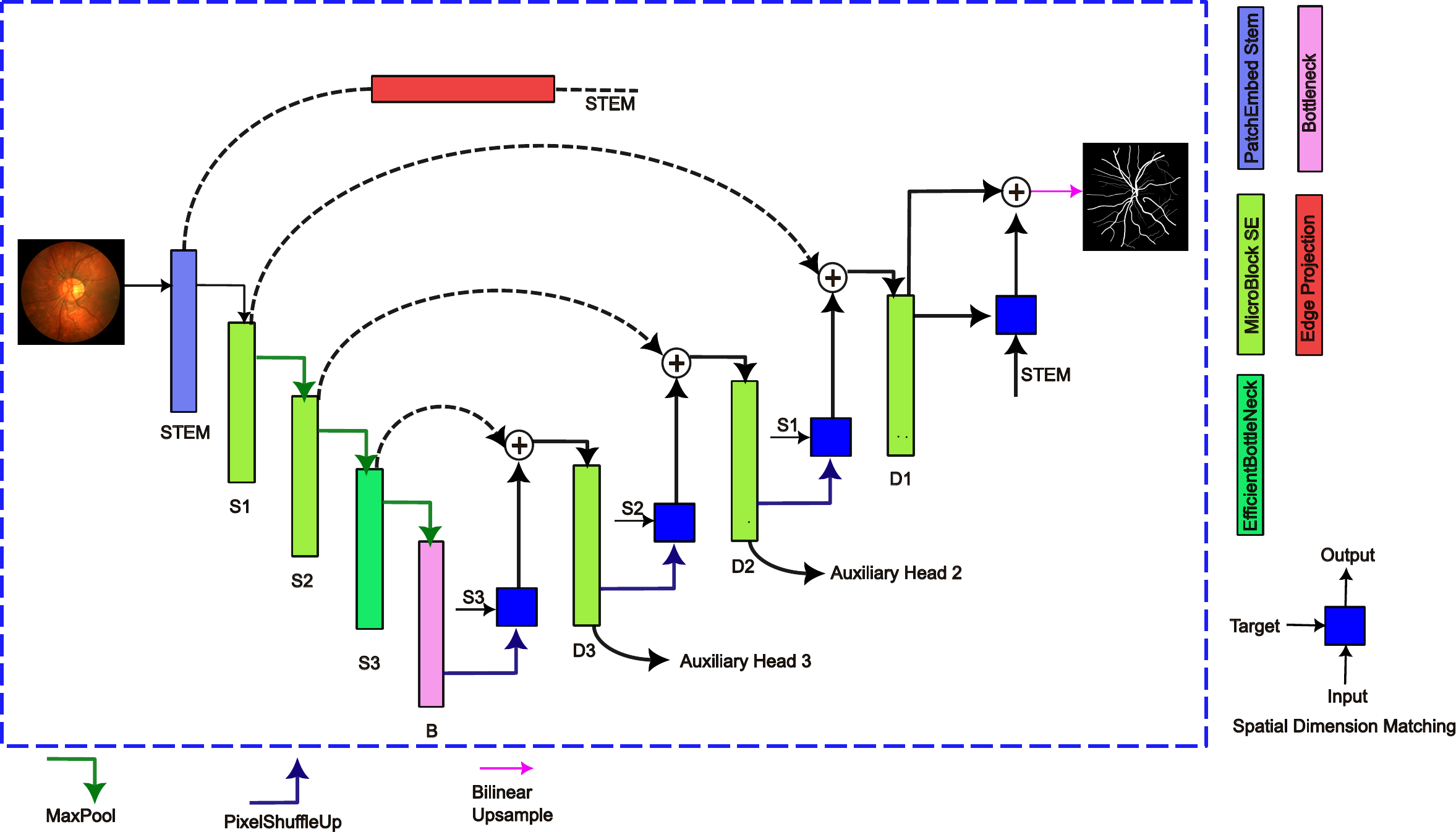}
    \caption{LightVesselNet Architecture}
    \label{fig:architecture}
\end{figure*}

The overall architecture follows a three-level encoder–bottleneck–decoder
structure, illustrated in Fig.~\ref{fig:architecture}.  An input fundus image of shape $H \times W \times 3$ is first received by a lightweight patch stem that downsamples the input by a factor of 2 but increases the channel number from 3 to 18. 

The encoder progressively extracts features and down-samples resolution through three stages; after each stage, a $2\times2$ max-pooling layer halves the spatial resolution and increases the effective receptive field of subsequent layers, allowing deeper stages to capture larger-scale vessel structures.

A dedicated multi-scale bottleneck aggregates features across four parallel dilation rates before a spatial attention gate focuses processing on the vessel-relevant regions. The decoder mirrors the encoder with three up-sampling stages; at each stage, a pixel-shuffle operation restores spatial resolution, and the result is concatenated with the corresponding encoder skip-connection feature map.  

A direct residual connection from the patch stem to the final decoder
output preserves fine-grained edge information.  The main segmentation logit
is produced by a $1 \times 1$ convolution after bilinear up-sampling back to
the original image resolution.  Two auxiliary output heads, attached at the
second and third decoder stages, provide deep supervision during training.

\begin{figure*}[!t]
\centering
\includegraphics[
    width=\textwidth,
    height=0.9\textheight,
    keepaspectratio
]{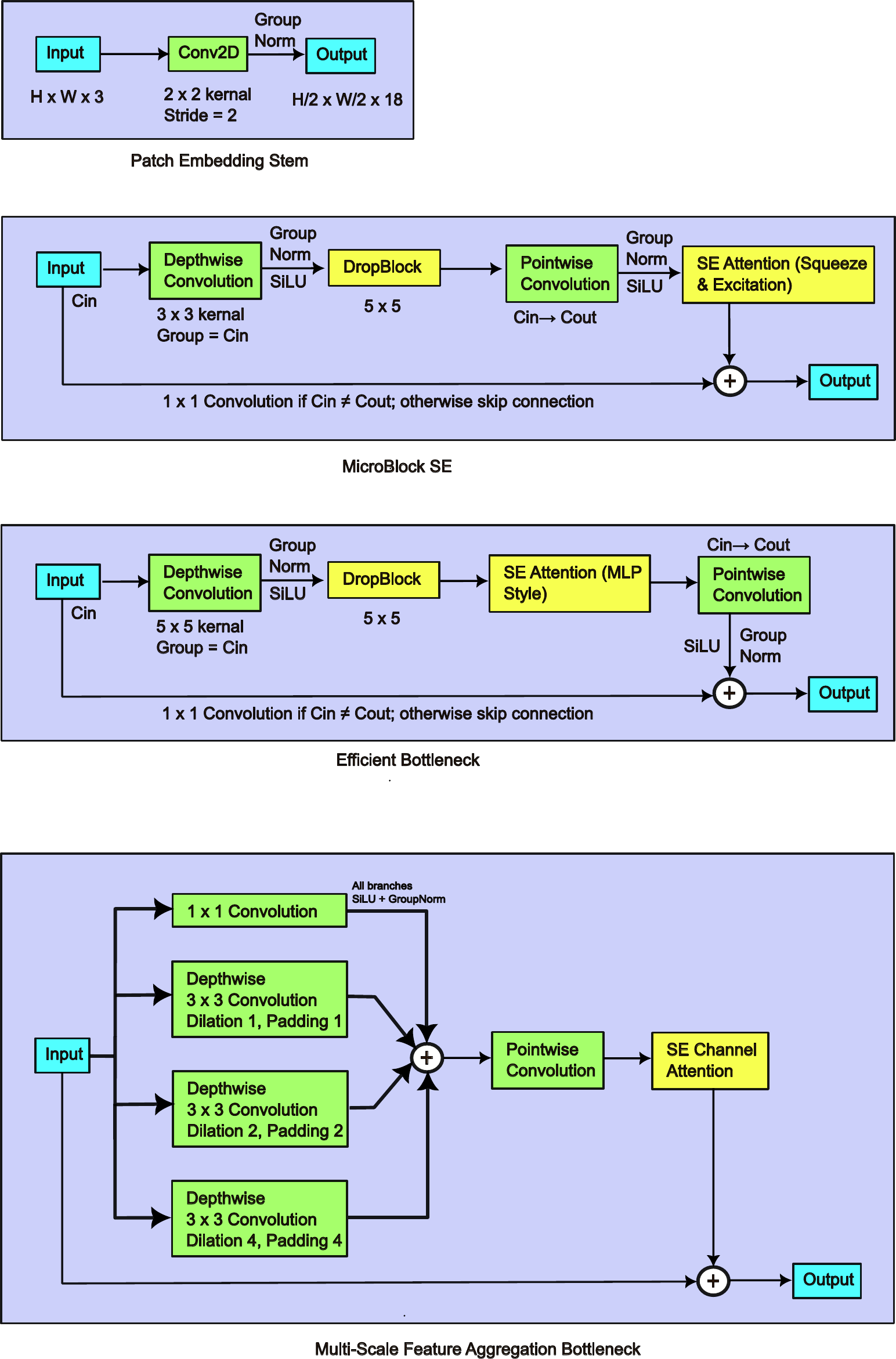}
\caption{Major Blocks used in LightVesselNet Architecture}
\label{fig:LVNet_Blocks}
\end{figure*}

Each component of the architecture is described in detail below. Their architectures are illustrated in Figure \ref{fig:LVNet_Blocks}.

\subsubsection{Patch Embedding Stem}

The network begins with a PatchEmbed module that replaces the common
practice of using two successive $3 \times 3$ convolutions for initial feature
extraction.  A single strided convolution with kernel size $2 \times 2$ and
stride $2$ maps the three-channel input to an 18-channel feature map at half the
spatial resolution ($H/2 \times W/2$).  Batch normalisation is replaced by
a Group Normalisation layer, which is more stable with small batch sizes (e.g., ~1).

By collapsing the first spatial down-sampling step into a single convolution,
the stem adds a negligible number of parameters ($3 \times 2 \times 2 \times 18 = 216$ weights) while immediately halving the memory footprint for all subsequent
operations.  The output of this stem is also retained as a \textit{high-resolution edge feature map} that is later injected into the final decoder stage via a projected residual connection, preserving low-level structural information that would otherwise be lost through the successive pooling operations of the encoder.

\subsubsection{MicroBlockSE}

The \textit{MicroBlockSE} module is the primary building block of the encoder
and decoder.  It replaces a standard $3 \times 3$ convolution (which has
$9 \times C_{\mathrm{in}} \times C_{\mathrm{out}}$ parameters) with a depthwise-separable convolution that has the same receptive field at a lower cost.  

Initially, a $3 \times 3$ depthwise convolution is applied to each input channel independently (groups $=C_{\mathrm{in}}$), capturing local spatial structure without mixing channels. The output is normalised via GroupNorm and passed through a SiLU (Sigmoid Linear Unit) activation, which has been shown to outperform ReLU in compact architectures owing to its smooth, non-monotonic shape.

A \textit{DropBlock} layer drops contiguous $5 \times 5$ blocks of activations during training rather than independent pixels, forcing the network to rely on spatially distributed cues.  This is particularly beneficial for retinal vessel segmentation, where vessels span extended spatial extents; dropping correlated spatial regions prevents the network from overfitting to local textural patterns and promotes learning of global vessel geometry.
A $1 \times 1$ convolution then projects the features to the desired output channel dimension $C_{\mathrm{out}}$, performing all cross-channel mixing.

A lightweight SE module recalibrates the output channels channel-wise.  Global average pooling compresses each feature map to a scalar, a two-layer bottleneck (with a compression ratio of~32) produces a channel-wise gate, and element-wise multiplication re-weights the feature maps.  This enables the block to emphasise channels that encode vessel-relevant information and suppress noisy background channels at negligible parameter cost.

The input is added back to the output via a $1 \times 1$ shortcut convolution (or an identity map when $C_{\mathrm{in}} = C_{\mathrm{out}}$), facilitating stable gradient flow.

The block can be expressed compactly as:
\begin{equation}
    \mathbf{y} = \mathrm{SE}\!\left(
        \mathrm{GN}\!\left( \mathrm{PW}\!\left(
            \mathrm{DropBlock}\!\left(
                \mathrm{GN}\!\left( \mathrm{DW}(\mathbf{x}) \right)
            \right)
        \right) \right)
    \right) + \mathrm{skip}(\mathbf{x}),
\end{equation}
where $\mathrm{DW}$ denotes $3 \times 3$ depthwise convolution, $\mathrm{PW}$
denotes $1 \times 1$ pointwise convolution, and $\mathrm{GN}$ denotes GroupNorm.

\textit{MicroBlockSE} is used at encoder stages Enc1 ($18 \to 26$ channels) and
Enc2 ($26 \to 40$ channels), and at all three decoder stages.

\subsubsection{EfficientBottleneck}

Encoder stage 3 uses an \textit{EfficientBottleneck} block, which differs
from MicroBlockSE in three aspects.  First, the depthwise convolution uses a
larger $5 \times 5$ kernel (padding $= 2$), providing a $5 \times 5$ receptive
field within the single block, and therefore better capturing medium-scale
vessel structure at the deepest encoder stage. Second, the positions of pointwise convolution and SE attention are interchanged. Third, the SE module is
implemented using two Linear layers operating on the spatially pooled representation, rather than $1 \times 1$ convolutions; this removes the
need for an extra reshape, and is marginally more memory-efficient at this stage, where the spatial dimensions are already small. This SE attention is different from the SE attention implemented in MicroBlock SE. 
The block maps $40 \to 60$ channels and accounts for a modest increase in parameter count relative to MicroBlockSE, justified by the richer feature representation needed at the deepest encoding level.

\subsubsection{Multi-Scale Feature Aggregation (MSFA) Bottleneck}
  
Retinal vessels span a wide range of diameters. A single fixed-kernel convolution favours one scale.  The \textit{Multi-Scale Feature Aggregation} (MSFA) bottleneck addresses this by processing the 60-channel bottleneck feature map with four parallel branches that operate simultaneously on different effective receptive fields. The 60-channel input is processed by four independent branches, each producing $q = 15$ feature channels (one quarter of 60):

\begin{itemize}
    \item \textbf{Branch 0 (local context):} A $1 \times 1$ pointwise
          convolution captures channel relationships without spatial mixing,
          serving as the identity-like reference.
    \item \textbf{Branch 1 (dilation 1):} A $3 \times 3$ depthwise convolution
          with dilation rate $d=1$ and padding $=1$, providing a standard
          $3 \times 3$ receptive field for fine-grained vessel detail.
    \item \textbf{Branch 2 (dilation 2):} A $3 \times 3$ depthwise convolution
          with $d=2$ and padding $=2$, giving an effective $5 \times 5$
          receptive field for medium-calibre vessels.
    \item \textbf{Branch 3 (dilation 4):} A $3 \times 3$ depthwise convolution
          with $d=4$ and padding $=4$, yielding an effective $9 \times 9$
          receptive field for larger vessels and spatial context.
\end{itemize}

All four branches apply GroupNorm and SiLU independently.  The branch outputs
are concatenated along the channel dimension to reconstruct the full 60-channel
width, followed by a $1 \times 1$ fusion convolution and channel-wise SE
attention.  A residual connection adds the original input back to the fused
output.

The MSFA can be written as:

\begin{equation}
\mathbf{y} = \mathrm{SE}\!\left( \mathrm{Fuse}\!\left(
\mathrm{Concat}\big( B_0(\mathbf{x}), B_1(\mathbf{x}), B_2(\mathbf{x}), B_3(\mathbf{x}) \big)
\right)\right) + \mathbf{x},
\end{equation}


Since all branches use \textit{depthwise} convolutions, the total parameter count of MSFA remains very low despite the four parallel paths.  

\subsubsection{Spatial Attention}

Immediately after the MSFA module, a lightweight \textit{Spatial Attention} gate is applied to the bottleneck features.  It computes channel-wise average
and maximum pooling across the feature dimension, and concatenates the two
$1 \times H_b \times W_b$ maps, and passes the result through a single
$7 \times 7$ convolution followed by a sigmoid activation.  The resulting attention map is multiplied element-wise with the input feature map:
\begin{equation}
    \mathbf{y} = \mathbf{x} \odot \sigma\!\left(
        f^{7 \times 7}\!\left( [\mathrm{AvgPool}(\mathbf{x}),\,
                                \mathrm{MaxPool}(\mathbf{x})] \right)
    \right).
\end{equation}
This attention mechanism guides the decoder to focus on vessel-bearing regions
of the image from the very beginning of the decoding path, suppressing
activations in the large background regions that dominate the fundus image.

\subsubsection{PixelShuffle Upsampling}

The decoder uses sub-pixel convolution (PixelShuffle) instead of transposed convolutions or bilinear interpolation followed by a convolution.  A $1 \times 1$ projection first expands the channel count by a factor of four ($C_{\mathrm{in}} \to 4C_{\mathrm{out}}$), then the PixelShuffle operator rearranges sub-pixel information into a $2 \times$ spatially up-sampled feature map of $C_{\mathrm{out}}$ channels without introducing the checkerboard artefacts associated with transposed convolutions.  GroupNorm is applied after the projection.

This design incurs zero additional parameters beyond the $1 \times 1$
projection and is computationally cheaper than a transposed convolution at
equivalent output resolution.

As retinal fundus images do not always possess spatial dimensions exactly divisible by the total encoder stride of 8, the decoder's upsampled feature maps may exhibit off-by-one spatial misalignments relative to their corresponding encoder skip connections. To avoid restricting the input resolution, we add a simple alignment step in the decoder. After each PixelShuffle upsampling stage, the feature map is padded only along the right and bottom edges using reflection padding so that it perfectly matches the skip connection size.

\subsubsection{Skip Connections and Edge Residual}

The decoder recovers fine spatial detail through two complementary skip
pathways. The standard encoder skip connections carry feature maps
$s_1$, $s_2$, and $s_3$ from the outputs of Enc1, Enc2, and Enc3,
respectively.  At each decoder stage, the up-sampled feature map is
concatenated with the corresponding skip feature before being processed by a
MicroBlockSE block ($120 \to 60$, $80 \to 40$, and $52 \to 26$ channels at
Dec3, Dec2, and Dec1, respectively).  Reflect padding is used to handle
off-by-one spatial mismatches that arise when the input height or width is not
exactly divisible by the total stride of $8$.

In addition to the encoder skip connections, a dedicated \textit{edge residual} path bypasses the entire encoder and bottleneck.  The stem output (18 channels,
$H/2 \times W/2$) is projected to 26 channels via a $1 \times 1$ convolution
followed by GroupNorm and SiLU, and added element-wise to the final Dec1
output.  This direct pathway injects the high-resolution, low-level edge, and
texture information that would otherwise be smoothed away by the three max-pooling steps and refocuses the decoder on sharp vessel boundaries.

\subsubsection{Output Heads}

The network has three prediction heads.  The \textbf{main head} is a
$1 \times 1$ convolution that maps the 26-channel Dec1 output to a single
logit channel, followed by bilinear up-sampling back to the full input
resolution $H \times W$.  During inference, a sigmoid activation converts the
logit to a probability map.

Two \textbf{auxiliary heads}, each a single $1 \times 1$ convolution, are
attached to the Dec3 and Dec2 outputs, producing lower-resolution
predictions at $H/4 \times W/4$ and $H/2 \times W/2$, respectively.  These
heads contribute to the composite training loss (Subsubsection \ref{subsub:loss}) with a weight of $w_\mathrm{aux} = 0.2$ and are discarded at inference time.  Deep supervision through auxiliary heads provides a more direct gradient signal to the early decoder layers, which would otherwise receive only heavily attenuated
gradients through the chain of MicroBlockSE blocks.

A summary of the network topology, channel widths, output resolutions, and layer-wise parameters is given in Table~\ref{tab:arch_summary}.

\begin{table}[!t]
\renewcommand{\arraystretch}{1.2}
\centering
\caption{LightVesselNet layer-by-layer summary (input: $H \times W \times 3$).}
\label{tab:arch_summary}

\resizebox{0.5\textwidth}{!}{%
\begin{tabular}{|l|l|c|c|c|}
\hline
\textbf{Stage} & \textbf{Module} & \textbf{Channels} & \textbf{Spatial size} & \textbf{Parameters} \\
\hline

Input          & --                                   & 3    & $H \times W$        & -      \\
Stem           & PatchEmbed (Conv $2\times2$, s2)     & 18   & $H/2 \times W/2$    & 252    \\
Enc1           & MicroBlockSE                         & 26   & $H/2 \times W/2$    & 1394   \\
Pool1          & MaxPool $2\times2$                   & 26   & $H/4 \times W/4$    & 0      \\
Enc2           & MicroBlockSE                         & 40   & $H/4 \times W/4$    & 2766   \\
Pool2          & MaxPool $2\times2$                   & 40   & $H/8 \times W/8$    & 0      \\
Enc3           & EfficientBottleneck                  & 60   & $H/8 \times W/8$    & 6320   \\
Pool3          & MaxPool $2\times2$                   & 60   & $H/16 \times W/16$  & 0      \\
Bottleneck     & MSFA + SpatialAttention             & 60   & $H/16 \times W/16$  & 6938   \\
Up3            & PixelShuffleUp                      & 60   & $H/8 \times W/8$    & 14880  \\
Dec3           & MicroBlockSE (cat $s_3$)            & $120 \rightarrow 60$ & $H/8 \times W/8$ & 16320  \\
Up2            & PixelShuffleUp                      & 40   & $H/4 \times W/4$    & 9920   \\
Dec2           & MicroBlockSE (cat $s_2$)            & $80 \rightarrow 40$  & $H/4 \times W/4$ & 7680   \\
Up1            & PixelShuffleUp                      & 26   & $H/2 \times W/2$    & 4368   \\
Dec1           & MicroBlockSE (cat $s_1$)            & $52 \rightarrow 26$  & $H/2 \times W/2$ & 3536   \\
Edge residual  & Conv $1\times1$ + add               & 26   & $H/2 \times W/2$    & 520    \\
Upsample       & Bilinear                            & 26   & $H \times W$        & 0     \\
Main out       & Conv $1\times1$                     & 1    & $H \times W$        & 129    \\
Aux head 3     & Conv $1\times1$                     & 1    & $H/8 \times W/8$    & 0      \\
Aux head 2     & Conv $1\times1$                     & 1    & $H/4 \times W/4$    & 0      \\
\hline
\end{tabular}%
}
\end{table}

\subsection{Implementation and Training Details}

\subsubsection{Pre-Processing and Data Augmentation}

The pre-processing techniques include green channel extraction from each RGB image and enhancing it by Contrast-Limited Adaptive Histogram Equalisation (CLAHE). The green channel offers the highest vessel-to-background contrast in colour fundus photography, and CLAHE is applied to further improve visibility by correcting uneven lighting and enhancing local contrast, while still avoiding unnecessary amplification of noise across the whole image.

Training samples were augmented using Albumentations. Spatial transforms included random flips, rotations, elastic deformations, grid distortions, and optical distortions. Photometric transforms comprised brightness-contrast jitter, gamma adjustment, unsharp masking, HSV jitter, Gaussian noise, Gaussian blur, and coarse dropout. All transforms were applied jointly to image--mask pairs to preserve spatial correspondence.  

\subsubsection{Loss Function}
\label{subsub:loss}

Training is supervised by a composite loss $\mathcal{L}$ that combines a
Tversky focal term with deep supervision:
 
\begin{equation}
  \mathcal{L} = \mathcal{L}_{\mathrm{TF}}(\hat{y}_{\mathrm{main}}, y)
    + w_{\mathrm{aux}}\!\left[
        \mathcal{L}_{\mathrm{TF}}(\hat{y}_{\mathrm{aux3}}, y)
      + \mathcal{L}_{\mathrm{TF}}(\hat{y}_{\mathrm{aux2}}, y)
    \right],
\end{equation}
 
where $\hat{y}_{\mathrm{main}}$ is the main decoder output,
$\hat{y}_{\mathrm{aux3}}$ and $\hat{y}_{\mathrm{aux2}}$ are auxiliary outputs
from the third and second decoder stages, respectively, $y$ is the ground-truth
mask, and $w_{\mathrm{aux}}=0.2$.
Auxiliary targets are resized to match the corresponding spatial resolution via
nearest-neighbour interpolation before loss computation.
 
The per-branch term $\mathcal{L}_{\mathrm{TF}}$ is:
 
\begin{equation}
  \mathcal{L}_{\mathrm{TF}}(\hat{y}, y) = \mathcal{L}_{\mathrm{Tversky}}(\hat{y}, y)
    + \mathcal{L}_{\mathrm{Focal}}(\hat{y}, y).
\end{equation}
 
\textbf{Tversky loss.}
The Tversky index generalises the Dice coefficient by independently weighting false negatives (FN) and false positives (FP):
 
\begin{equation}
  \mathcal{L}_{\mathrm{Tversky}} = 1 -
  \frac{\mathrm{TP} + \varepsilon}
       {\mathrm{TP} + \alpha\,\mathrm{FN} + \beta\,\mathrm{FP} + \varepsilon},
\end{equation}
 
where $\alpha$ is a hyperparameter, $\beta = 1 - \alpha$, and $\varepsilon=10^{-6}$.

\textbf{Focal BCE loss.}
The focal loss down-weights well-classified background pixels and focuses training on hard vessel pixels:
 
\begin{equation}
  \mathcal{L}_{\mathrm{Focal}} =
    \mathbb{E}\!\left[
      \left(1 - e^{-\ell_{\mathrm{BCE}}}\right)^{\!\gamma} \cdot \ell_{\mathrm{BCE}}
    \right],
\end{equation}

where $\ell_{\mathrm{BCE}}$ is the binary cross-entropy computed from the
raw logits and $\gamma$ is the focusing parameter.

\subsubsection{Experimental Setup}

Training was done on Kaggle using NVIDIA T-4 GPU. The AdamW optimiser was used with an initial learning rate of $8\times10^{-3}$ and weight decay of $5\times10^{-4}$. The learning rate was scheduled using a one-cycle cosine
policy. The model checkpoint was saved whenever the composite score $\frac{1}{2}(\text{F1}+\text{SE})$ on the validation split improved.  

The number of training epochs, patience, and learning rate are set to 500, 50, and $8 \times 10^{-3}$, respectively. For loss function hyperparameters, $\alpha$ and $\gamma$ are 0.5 and 2.

\subsection{Dataset Description \& Model Training}
\label{subsec:Dataset}
The datasets used in this study, along with the training method taken, are described in Table \ref{tab:dataset_overview}:

\begin{table}[h]
\caption{Overview of retinal vessel segmentation datasets used in this study along with Training Methods.}
\label{tab:dataset_overview}
\centering
\renewcommand{\arraystretch}{1.15}
\setlength{\tabcolsep}{5pt}
\resizebox{0.5\textwidth}{!}{%
\begin{tabular}{|l|c|c|c|c|}
\hline
\textbf{Dataset}                         & \textbf{Images} & \textbf{Train/Test} & \textbf{Resolution}                                                                         & \textbf{Evaluation Protocol} \\ \hline
DRIVE \cite{bib18}      & 40              & 20 / 20             & $565 \times 584$                                                                            & Standard training            \\ \hline
STARE \cite{bib17}      & 20              & --                  & $700 \times 605$                                                                            & Leave-one-out validation     \\ \hline
CHASE\_DB1 \cite{bib19} & 28              & --                  & $999 \times 960$                                                                            & K-fold cross-validation      \\ \hline
FIVES \cite{bib2}       & 800             & 600 / 200           & \begin{tabular}[c]{@{}c@{}}$2048 \times 2048$\\ Resized to \\ $512 \times 512 $\end{tabular}   & Standard Training            \\ \hline
HRF \cite{bib26}        & 45              & -                   & \begin{tabular}[c]{@{}c@{}}$3504 \times 2336$\\ Resized to \\ $1752 \times 1168 $\end{tabular} & K-fold cross-validation      \\ \hline
\end{tabular}
}
\end{table}

As the \textbf{DRIVE} and \textbf{FIVES} datasets are officially split into a train and a test set, there was no chance for test data leakage. However, for the remaining three datasets, an official split was not available. As \textbf{STARE} has only 20 images, we used the leave-one-out validation method, where each image was used for testing once. In each fold, 18 images were used for training, 1 for validation, and the remaining 1 for testing. In this way, training was done for 20 folds so that each image is evaluated once. 

The \textbf{CHASE\_DB1} dataset contains higher-resolution images and a larger number of images compared to the STARE dataset. So, to fasten the training process, we applied K-fold cross-validation, where K = 7. In each fold, 4 images were in the test set, 2 for validation, and 22 images for training. So, in 7 folds, all images were tested once. Similarly, the \textbf{HRF} dataset has set an even higher resolution with no official test-train split. So, a 5-fold cross-validation was applied. For STARE, CHASE\_DB1, and HRF datasets, the final results were found by averaging the results across all folds.

As the input resolution was very high for the FIVES and HRF datasets, and LightVesselNet is explicitly designed for resource-constrained edge devices, we resized the images of these datasets. For the remaining datasets, resizing was not applied.

\section{Experiments and Results}
\label{sec:Experiments}

\subsection{Computational Efficiency Analysis}
To evaluate whether the proposed model is lightweight and deployable on edge devices, we considered several metrics. Table \ref{tab:Comp_Eff} shows the value of different metrics used to investigate computational efficiency. 

Among these metrics, the number of parameters and model size are independent of input size. For our model, there are no non-trainable parameters, and so the total number of parameters equals the total number of trainable parameters. Contrary to these two parameters, the remaining considered metrics depend on the input size. Thus, we regarded various input sizes, such as common resolutions like $256 \times 256$, and common datasets' input resolution.

\begin{table*}[!t]
\centering
\caption{Complexity and Runtime Breakdown}
\label{tab:Comp_Eff}
\begin{tabular}{|l|c|c|c|c|c|c|c|c|}
\hline
\textbf{Image Size} & \textbf{Params (M)} & \textbf{Model Size} & \textbf{GFLOPs} & \textbf{FPS (GPU)} & \textbf{GPU Time (ms)} & \textbf{FPS (CPU)} & \textbf{CPU Time (ms)} & \textbf{Peak GPU MB} \\ \hline
256 $\times$ 256       & \multirow{8}{*}{0.0750}        & \multirow{8}{*}{0.2862 MB}     & 0.3510   & 206.79     & 4.84         & 26.96      & 37.09          & 70.4   \\
512 $\times$ 512       &                                &                                & 1.4041   & 128.28     & 7.80         & 7.82       & 127.84         & 115.6  \\
584 $\times$ 565       &                                &                                & 1.7609   & 106.52     & 9.39         & 5.27       & 189.82         & 133.5  \\
700 $\times$ 605       &                                &                                & 2.2572   & 81.92      & 12.21        & 3.54       & 282.51         & 152.4  \\
768 $\times$ 768       &                                &                                & 3.1593   & 59.91      & 16.69        & 2.20       & 454.45         & 191.8  \\
999 $\times$ 960       &                                &                                & 5.1221   & 36.33      & 27.52        & 1.74       & 574.00         & 278.3  \\
1024 $\times$ 1024     &                                &                                & 5.6165   & 33.93      & 29.48        & 1.64       & 611.12         & 297.9  \\
1752 $\times$ 1158     &                                &                                & 10.8477  & 16.75      & 59.71        & 0.52       & 1934.36        & 528.0  \\
\hline
\end{tabular}
\end{table*}

The analysis confirms that the proposed model achieves a substantially reduced computational footprint, comprising only 75K parameters and modest GFLOPs, while sustaining competitive inference throughput across a range of input resolutions. These results demonstrate that the proposed model achieves effective feature learning with minimal trainable parameters and arithmetic operations. Furthermore, the model produces predictions at low latency, as evidenced by the GPU and CPU inference times reported in the Table.  Overall, the proposed model is suitable for deployment in resource-constrained edge environments.

\subsection{Performance on Individual Datasets}
LightVesselNet's performance on each dataset is shown in Table \ref{tab:res}. The table also contains the formulas for different metrics used. Given the approximate 1:10 vessel-to-background pixel ratio characteristic of retinal fundus images, Sensitivity (true positive rate) is a particularly critical evaluation metric for this task, as it directly reflects the model's ability to detect vessel pixels.

                         






\begin{table*}[!t]
\centering
\caption{Result}
\label{tab:res}
\renewcommand{\arraystretch}{1.6}
\resizebox{\textwidth}{!}{%
\begin{tabular}{|c|c|c|c|c|c|c|c|}
\hline

\multirow{2}{*}{Dataset} 
& Sensitivity 
& Specificity 
& Accuracy 
& AUC 
& Precision 
& Dice / F1
& Jaccard / IoU \\ 

& $\frac{\mathrm{TP}}{\mathrm{TP+FN}}$ 
& $\frac{\mathrm{TN}}{\mathrm{TN+FP}}$ 
& $\frac{\mathrm{TP+TN}}{\mathrm{TP+TN+FP+FN}}$ 
& $\int \mathrm{TPR}\, d\mathrm{FPR}$ 
& $\frac{\mathrm{TP}}{\mathrm{TP+FP}}$ 
& $\frac{2\mathrm{TP}}{2\mathrm{TP+FP+FN}}$ 
& $\frac{\mathrm{TP}}{\mathrm{TP+FP+FN}}$ \\

\hline

DRIVE      & 0.8189$\pm$0.0060 & 0.9803$\pm$0.0041 & 0.9659$\pm$0.0022 & 0.9830$\pm$0.0042 & 0.8957$\pm$0.0134 & 0.8070$\pm$0.0052 & 0.6766$\pm$0.0048 \\

STARE      & 0.8499$\pm$0.0026 & 0.9799$\pm$0.0020 & 0.9696$\pm$0.0013 & 0.9878$\pm$0.0016 & 0.9011$\pm$0.0049 & 0.8072$\pm$0.0019 & 0.6779$\pm$0.0040 \\

CHASE\_DB1 & 0.8640$\pm$0.0024 & 0.9821$\pm$0.0043 & 0.9738$\pm$0.0033 & 0.9896$\pm$0.0033 & 0.9072$\pm$0.0035 & 0.8181$\pm$0.0091 & 0.6933$\pm$0.0012 \\

FIVES      & 0.8634$\pm$0.0069 & 0.9912$\pm$0.0029 & 0.9830$\pm$0.0054 & 0.9907$\pm$0.0025 & 0.9352$\pm$0.0099 & 0.8649$\pm$0.0054 & 0.7693$\pm$0.0023 \\

HRF        & 0.8096$\pm$0.0043 & 0.9766$\pm$0.0042 & 0.9634$\pm$0.0081 & 0.9763$\pm$0.0033 & 0.8557$\pm$0.0065 & 0.7686$\pm$0.0057 & 0.6270$\pm$0.0012 \\

\hline
\end{tabular}%
}
\end{table*}

The training technique used for each dataset is different and explained in detail in Subsection \ref{subsec:Dataset}. 

\begin{figure*}[ht]
    \centering
    \includegraphics[width=0.6\textwidth]{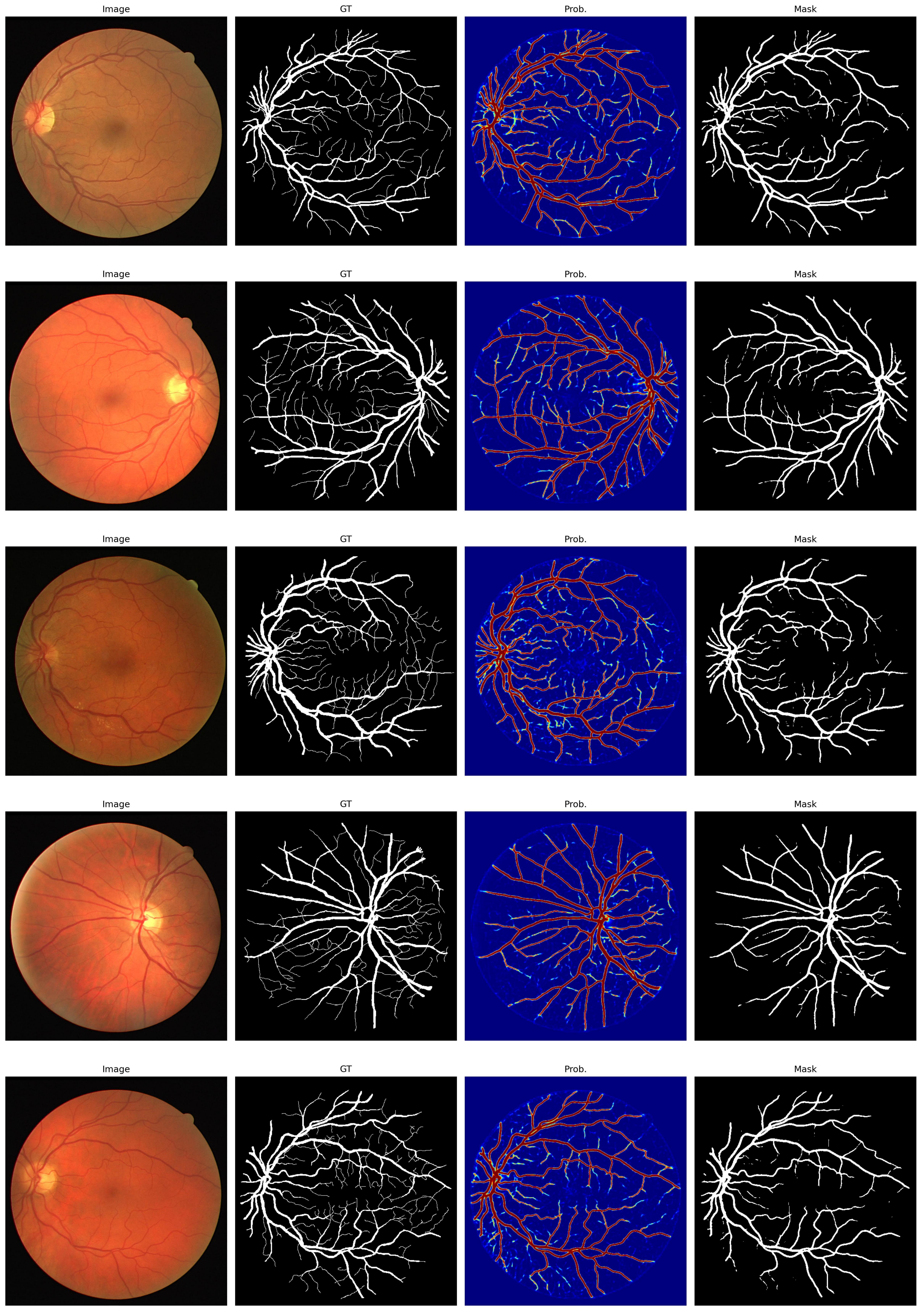}
    \caption{Qualitative segmentation results produced by LightVesselNet. From left to right: input image, ground truth, probability map(blue means negative and red means positive), predicted mask.}
    \label{fig:Visualization}
\end{figure*}

Figure \ref{fig:Visualization} illustrates some fundus images, their respective ground truths, probability maps, and LightVesselNet's predicted masks.

\subsection{Cross-Dataset Evaluation}
To demonstrate the generalisation capability of LightVesselNet, we conducted a cross-dataset evaluation where the model was trained on a full dataset and tested on the remaining four datasets. The result of this experiment is shown in Table \ref{tab:CrossData}. 

\begin{table*}[!t]
\centering
\caption{Cross-Dataset Evaluation}
\label{tab:CrossData}
\renewcommand{\arraystretch}{1.6}
\resizebox{\textwidth}{!}
{%
\begin{tabular}{|c|ccccccccccccccc|}
\hline
\multirow{3}{*}{Trained On} & \multicolumn{15}{c|}{Tested On}                                                                                                                                                                                                                                                                                                                                                                                                             \\ \cline{2-16} 
                            & \multicolumn{3}{c|}{DRIVE}                                                              & \multicolumn{3}{c|}{STARE}                                                              & \multicolumn{3}{c|}{Chase\_DB1}                                                         & \multicolumn{3}{c|}{FIVES}                                                               & \multicolumn{3}{c|}{HRF}                                           \\ \cline{2-16}
                 & \multicolumn{1}{c|}{Se}     & \multicolumn{1}{c|}{Sp}     & \multicolumn{1}{c|}{Acc}    & \multicolumn{1}{c|}{Se}     & \multicolumn{1}{c|}{Sp}     & \multicolumn{1}{c|}{Acc}    & \multicolumn{1}{c|}{Se}     & \multicolumn{1}{c|}{Sp}     & \multicolumn{1}{c|}{Acc}    & \multicolumn{1}{c|}{Se}     & \multicolumn{1}{c|}{Sp}      & \multicolumn{1}{c|}{Acc}    & \multicolumn{1}{c|}{Se}     & \multicolumn{1}{c|}{Sp}     & Acc    \\ \hline
DRIVE                       & \multicolumn{3}{c|}{-}                                                                  & \multicolumn{1}{l|}{0.8517} & \multicolumn{1}{l|}{0.9742} & \multicolumn{1}{l|}{0.9645} & \multicolumn{1}{l|}{0.8349} & \multicolumn{1}{l|}{0.9724} & \multicolumn{1}{l|}{0.9626} & \multicolumn{1}{l|}{0.8413} & \multicolumn{1}{l|}{0.9879}  & \multicolumn{1}{l|}{0.9774} & \multicolumn{1}{l|}{0.7402} & \multicolumn{1}{l|}{0.9621} & 0.9446 \\ \hline
STARE                       & \multicolumn{1}{l|}{0.7662} & \multicolumn{1}{l|}{0.9789} & \multicolumn{1}{l|}{0.9601} & \multicolumn{3}{c|}{-}                                                                  & \multicolumn{1}{l|}{0.8641} & \multicolumn{1}{l|}{0.9672} & \multicolumn{1}{l|}{0.9600} & \multicolumn{1}{l|}{0.7783} & \multicolumn{1}{l|}{0.9873}  & \multicolumn{1}{l|}{0.9702} & \multicolumn{1}{l|}{0.7322} & \multicolumn{1}{l|}{0.9665} & 0.9442 \\ \hline
CHASE\_DB1                  & \multicolumn{1}{l|}{0.8087} & \multicolumn{1}{l|}{0.9710} & \multicolumn{1}{l|}{0.9566} & \multicolumn{1}{l|}{0.8322} & \multicolumn{1}{l|}{0.9691} & \multicolumn{1}{l|}{0.9583} & \multicolumn{3}{c|}{-}                                                                  & \multicolumn{1}{l|}{0.8510} & \multicolumn{1}{l|}{0.97490} & \multicolumn{1}{l|}{0.9657} & \multicolumn{1}{l|}{0.7318} & \multicolumn{1}{l|}{0.9568} & 0.9390 \\ \hline
FIVES                       & \multicolumn{1}{l|}{0.7638} & \multicolumn{1}{l|}{0.9813} & \multicolumn{1}{l|}{0.9594} & \multicolumn{1}{l|}{0.8447} & \multicolumn{1}{l|}{0.9726} & \multicolumn{1}{l|}{0.9626} & \multicolumn{1}{l|}{0.8455} & \multicolumn{1}{l|}{0.9735} & \multicolumn{1}{l|}{0.9644} & \multicolumn{3}{c|}{-}                                                                   & \multicolumn{1}{l|}{0.7862} & \multicolumn{1}{l|}{0.9592} & 0.9456 \\ \hline
HRF                         & \multicolumn{1}{l|}{0.7750} & \multicolumn{1}{l|}{0.9712} & \multicolumn{1}{l|}{0.9538} & \multicolumn{1}{l|}{0.8538} & \multicolumn{1}{l|}{0.9637} & \multicolumn{1}{l|}{0.9551} & \multicolumn{1}{l|}{0.8585} & \multicolumn{1}{l|}{0.9609} & \multicolumn{1}{l|}{0.9536} & \multicolumn{1}{l|}{0.8617} & \multicolumn{1}{l|}{0.9797}  & \multicolumn{1}{l|}{0.9718} & \multicolumn{3}{c|}{-}                                             \\ \hline
\end{tabular}
}
\end{table*}

Cross-dataset evaluation requires a uniform input resolution to ensure that performance differences reflect true domain generalisation rather than resolution variation; $512 \times 512$ was therefore adopted as the common dimension. Resizing high-resolution datasets such as FIVES and HRF introduces some loss of fine spatial detail, so cross-dataset results should be interpreted as indicators of domain adaptability rather than absolute performance at native resolution.

The results in Table~\ref{tab:CrossData} demonstrate that LightVesselNet retains reasonable segmentation capability on unseen domains without fine-tuning. Several source-target combinations yield strong performance: models trained on DRIVE achieve sensitivities of 0.8517 and 0.8349 on STARE and CHASE\_DB1, respectively, while HRF-trained models reach 0.8538, 0.8585, and 0.8617 on STARE, CHASE\_DB1, and FIVES — closely approaching within-dataset levels reported in Table \ref{tab:res}. 

Performance reduction in certain source-target pairs is consistent with well-documented domain shift challenges in retinal vessel segmentation \cite{bib11}, and reflects a field-wide limitation rather than a model-specific one.

To our knowledge, none of the lightweight comparative models — including LFA-Net, LFRA-Net, and RetinaLiteNet — report cross-dataset results. LightVesselNet is therefore the first sub-100K parameter retinal vessel segmentation model to be systematically evaluated across five benchmark datasets, providing a reference baseline for future work and highlighting cross-domain evaluation as an important standard reporting practice.

\subsection{Ablation Study}
We conduct an ablation study to assess the contribution of each module in the proposed architecture. The result is shown in Table \ref{tab:model_ablation}

\begin{table*}[!t]
\centering
\caption{Model component ablation on DRIVE and STARE. Best result per metric in \textbf{bold}.}
\label{tab:model_ablation}
\setlength{\tabcolsep}{4pt}
\resizebox{\textwidth}{!}
{%
\begin{tabular}{|l|l|cccc|cccc|}
\hline
\multirow{2}{*}{Axis}                & \multirow{2}{*}{Variant}                                   & \multicolumn{4}{c|}{DRIVE}                                                                                                           & \multicolumn{4}{c|}{STARE}                                                                                                           \\ \cline{3-10} 
                                     &                                                            & \multicolumn{1}{c|}{SE}              & \multicolumn{1}{c|}{SP}              & \multicolumn{1}{c|}{F1}              & IoU             & \multicolumn{1}{c|}{SE}              & \multicolumn{1}{c|}{SP}              & \multicolumn{1}{c|}{F1}              & IoU             \\ \hline
Proposed model                       & all components & \multicolumn{1}{c|}{\textbf{0.8189}} & \multicolumn{1}{c|}{0.9803}          & \multicolumn{1}{c|}{\textbf{0.8070}} & \textbf{0.6766} & \multicolumn{1}{c|}{\textbf{0.8499}} & \multicolumn{1}{c|}{\textbf{0.9799}} & \multicolumn{1}{c|}{\textbf{0.8072}} & \textbf{0.6779} \\ \hline
\multirow{3}{*}{Attention}           & w/o Attention                                              & \multicolumn{1}{c|}{0.7980}          & \multicolumn{1}{c|}{0.9762}          & \multicolumn{1}{c|}{0.7866}          & 0.6549          & \multicolumn{1}{c|}{0.8175}          & \multicolumn{1}{c|}{0.9743}          & \multicolumn{1}{c|}{0.7869}          & 0.6556          \\ \cline{2-10} 
                                     & w/o SE Only                                                & \multicolumn{1}{c|}{0.7964}          & \multicolumn{1}{c|}{\textbf{0.9810}} & \multicolumn{1}{c|}{0.7902}          & 0.6587          & \multicolumn{1}{c|}{0.8204}          & \multicolumn{1}{c|}{0.9764}          & \multicolumn{1}{c|}{0.7935}          & 0.6591          \\ \cline{2-10} 
                                     & w/o SA Only                                                & \multicolumn{1}{c|}{0.8015}          & \multicolumn{1}{c|}{0.9768}          & \multicolumn{1}{c|}{0.7888}          & 0.6524          & \multicolumn{1}{c|}{0.8285}          & \multicolumn{1}{c|}{0.9774}          & \multicolumn{1}{c|}{0.7891}          & 0.6570          \\ \hline
Bottleneck                           & w/o MSFA                                                   & \multicolumn{1}{c|}{0.7653}          & \multicolumn{1}{c|}{0.9659}          & \multicolumn{1}{c|}{0.7578}          & 0.6358          & \multicolumn{1}{c|}{0.7932}          & \multicolumn{1}{c|}{0.9665}          & \multicolumn{1}{c|}{0.7571}          & 0.6385          \\ \hline
Edge residual                        & w/o Edge Skip                                              & \multicolumn{1}{c|}{0.7914}          & \multicolumn{1}{c|}{0.9766}          & \multicolumn{1}{c|}{0.7876}          & 0.6596          & \multicolumn{1}{c|}{0.8205}          & \multicolumn{1}{c|}{0.9762}          & \multicolumn{1}{c|}{0.7874}          & 0.6603          \\ \hline
\multirow{2}{*}{Upsampling Strategy} & Transposed Convolution                                     & \multicolumn{1}{c|}{0.8092}          & \multicolumn{1}{c|}{0.9804}          & \multicolumn{1}{c|}{0.7940}          & 0.6568          & \multicolumn{1}{c|}{0.8287}          & \multicolumn{1}{c|}{0.9794}          & \multicolumn{1}{c|}{0.7958}          & 0.6467          \\ \cline{2-10} 
                                     & Bilinear Interpolation                                     & \multicolumn{1}{c|}{0.7943}          & \multicolumn{1}{c|}{0.9801}          & \multicolumn{1}{c|}{0.7756}          & 0.6426          & \multicolumn{1}{c|}{0.81285}         & \multicolumn{1}{c|}{0.9789}          & \multicolumn{1}{c|}{0.7845}          & 0.6436          \\ \hline
Normalization                        & GroupNorm                                                  & \multicolumn{1}{c|}{0.8189}          & \multicolumn{1}{c|}{0.9803}          & \multicolumn{1}{c|}{0.8070}          & 0.6766          & \multicolumn{1}{c|}{0.8499}          & \multicolumn{1}{c|}{0.9799}          & \multicolumn{1}{c|}{0.8072}          & 0.6779          \\ \hline
Deep Supervision                     & w/o Deep Supervision                                       & \multicolumn{1}{c|}{0.8115}          & \multicolumn{1}{c|}{0.9801}          & \multicolumn{1}{c|}{0.8023}          & 0.6687          & \multicolumn{1}{c|}{0.8421}          & \multicolumn{1}{c|}{0.9717}          & \multicolumn{1}{c|}{0.8006}          & 0.6737          \\ \hline
\multirow{4}{*}{Regularization}      & No Regularization                                          & \multicolumn{1}{c|}{0.7652}          & \multicolumn{1}{c|}{0.9804}          & \multicolumn{1}{c|}{0.7631}          & 0.6378          & \multicolumn{1}{c|}{0.8159}          & \multicolumn{1}{c|}{0.9754}          & \multicolumn{1}{c|}{0.7989}          & 0.6637          \\ \cline{2-10} 
                                     & Dropout                                                    & \multicolumn{1}{c|}{0.7986}          & \multicolumn{1}{c|}{0.9801}          & \multicolumn{1}{c|}{0.7836}          & 0.6648          & \multicolumn{1}{c|}{0.8276}          & \multicolumn{1}{c|}{0.9787}          & \multicolumn{1}{c|}{0.7948}          & 0.6721          \\ \cline{2-10} 
                                     & $3 \times 3$ DropBlock                                     & \multicolumn{1}{c|}{0.8143}          & \multicolumn{1}{c|}{0.9789}          & \multicolumn{1}{c|}{0.8051}          & 0.6734          & \multicolumn{1}{c|}{0.8431}          & \multicolumn{1}{c|}{0.9776}          & \multicolumn{1}{c|}{0.8044}          & 0.6761          \\ \cline{2-10} 
                                     & $7 \times 7$ DropBlock                                     & \multicolumn{1}{c|}{0.8155}          & \multicolumn{1}{c|}{0.9806}          & \multicolumn{1}{c|}{0.8065}          & 0.6726          & \multicolumn{1}{c|}{0.8458}          & \multicolumn{1}{c|}{0.9792}          & \multicolumn{1}{c|}{0.8053}          & 0.6772          \\ \hline
\end{tabular}
}
\end{table*}

Notably, the largest reductions are observed in sensitivity when MSFA is removed, suggesting that MSFA is particularly important for detecting thin and low-contrast vessels. In contrast, specificity remains relatively stable across most variants, indicating that the improvements primarily arise from enhanced vessel recovery rather than increased background suppression. All the variants have almost similar parameter count; thus, we choose the version with the best performance.

\subsection{Comparison with State-of-the-Art Methods}

We compare the proposed LightVesselNet with several state-of-the-art methods on benchmark datasets, using standard evaluation metrics to assess its performance. Most of the works measured their performance on DRIVE, STARE, and CHASE\_DB1. As a result, there were few works to compare in the HRF and FIVES datasets. So, we had to compare even with heavier models in these datasets. The comparison is presented in Table \ref{tab:drive_comparison}, \ref{tab:stare_comparison}, \ref{tab:chasedb1_comparison}, \ref{tab:hrf_comparison}, \ref{tab:fives_comparison}.

\begin{table*}[t]
\centering
\caption{Comparison of Lightweight Retinal Vessel Segmentation Methods on DRIVE Dataset}
\label{tab:drive_comparison}
\renewcommand{\arraystretch}{1.3}
\begin{tabular}{|l|c|c|c|c|c|c|}
\hline
\textbf{Work} & \textbf{Params} & \textbf{SE} & \textbf{SP} & \textbf{ACC} & \textbf{F1} & \textbf{Jaccard} \\
 & \textbf{(M)} & & & & & \\
\hline
UNet (2015) \cite{bib13}                     & 31.03 & 0.7727 & 0.9866 & 0.9675 & 0.8040  & 0.6801 \\
RetinaLiteNet (2024) \cite{bib20} & \textbf{0.066} & 0.7840 & 0.9800 & --     & 0.8060 & 0.6750 \\
LVS-Net (2024) \cite{bib21}           & 0.71  & \textbf{0.8391} & 0.9851 & \textbf{0.9664} & \textbf{0.8644} & \textbf{0.7616} \\
LFA-Net (2025) \cite{bib22}           & 0.11  & 0.8056 & 0.9809 & 0.9609 & 0.8318 & 0.7124 \\
LFRA-Net (2025) \cite{bib23}          & 0.17  & 0.8243 & 0.9808 & 0.9609 & 0.8428 & 0.7286 \\
LW U-Net + RA (2025) \cite{bib24}     & 1.94  & 0.7421 & \textbf{0.9837} & 0.9113 & 0.7871 & 0.6318 \\
\hline
\textbf{LightVesselNet (Ours)}        & \textbf{0.075} & 0.8189 & 0.9803 & 0.9659 & 0.8070 & 0.6766 \\
\hline
\end{tabular}%
\end{table*}

\begin{table*}[t]
\centering
\caption{Comparison of Lightweight Retinal Vessel Segmentation Methods on STARE Dataset}
\label{tab:stare_comparison}
\renewcommand{\arraystretch}{1.3}
\begin{tabular}{|l|c|c|c|c|c|c|}
\hline
\textbf{Work} & \textbf{Params} & \textbf{SE} & \textbf{SP} & \textbf{ACC} & \textbf{F1} & \textbf{Jaccard} \\
 & \textbf{(M)} & & & & & \\
\hline
UNet (2015) \cite{bib13}                     & 31.03 & 0.8012 & 0.9765 & 0.9624 & 0.7915 & 0.6887 \\
LVS-Net (2024) \cite{bib21}           & 0.71  & 0.8740 & 0.9860 & \textbf{0.9739} & 0.8788 & 0.7840 \\
LFA-Net (2025) \cite{bib22}           & 0.11  & 0.8813 & 0.9830 & --     & 0.8716 & 0.7729 \\
LFRA-Net (2025) \cite{bib23}          & 0.17  & \textbf{0.8875} & \textbf{0.9856} & --     & \textbf{0.8844} & \textbf{0.7931} \\
\hline
\textbf{LightVesselNet (Ours)}        & \textbf{0.075} & 0.8499 & 0.9799 & 0.9696 & 0.8072 & 0.6779 \\
\hline
\end{tabular}%
\end{table*}

\begin{table*}[t]
\centering
\caption{Comparison of Lightweight Retinal Vessel Segmentation Methods on CHASE\_DB1 Dataset}
\label{tab:chasedb1_comparison}
\renewcommand{\arraystretch}{1.3}
\begin{tabular}{|l|c|c|c|c|c|c|}
\hline
\textbf{Work} & \textbf{Params} & \textbf{SE} & \textbf{SP} & \textbf{ACC} & \textbf{F1} & \textbf{Jaccard} \\
 & \textbf{(M)} & & & & & \\
\hline
UNet (2015)\cite{bib13}                     & 31.03 & 0.8587 & 0.9866 & 0.9772 & 0.8407  & 0.7255 \\
LVS-Net (2024) \cite{bib21}           & 0.71  & 0.8343 & 0.9819 & 0.9644 & 0.8478 & 0.7365 \\
LFA-Net (2025) \cite{bib22}           & 0.11  & 0.8225 & 0.9815 & --     & 0.8405 & 0.7252 \\
LFRA-Net (2025) \cite{bib23}          & 0.17  & 0.8436 & 0.9820 & --     & \textbf{0.8550} & \textbf{0.7470} \\
LW U-Net + RA (2025) \cite{bib24}     & 1.94  & 0.8220 & \textbf{0.9843} & 0.9718 & 0.7946 & 0.6910 \\
\hline
\textbf{LightVesselNet (Ours)}        & \textbf{0.075} & \textbf{0.8640} & 0.9821 & \textbf{0.9738} & 0.8181 & 0.6933 \\
\hline
\end{tabular}%
\end{table*}

\begin{table*}[t]
\centering
\caption{Comparison of Lightweight Retinal Vessel Segmentation Methods on HRF Dataset}
\label{tab:hrf_comparison}
\renewcommand{\arraystretch}{1.3}
\begin{tabular}{|l|c|c|c|c|c|c|}
\hline
\textbf{Work} & \textbf{Params} & \textbf{SE} & \textbf{SP} & \textbf{ACC} & \textbf{F1 / Dice} & \textbf{Jaccard / IoU} \\
 & \textbf{(M)} & & & & & \\
\hline
UNet (2015) \cite{bib13}                     & 31.03 & 0.8062 & 0.9779 & 0.9608 & 0.7564 & 0.6396 \\
LHU-VT (2025) \cite{bib40}                               & 19.40 & \textbf{0.8321} & \textbf{0.9851} & \textbf{0.9659} & \textbf{0.8598} & \textbf{0.7540} \\
LW U-Net + RA (2025) \cite{bib24}                         & 1.94  & 0.8161 & 0.9707 & 0.8437 & 0.6902 & 0.5270 \\
GraphSeg (2025)\cite{bib41}                              & 19.32 & 0.8158 & 0.9807 & 0.9588 & 0.8389 & --     \\
\hline
\textbf{LightVesselNet (Ours)}        & \textbf{0.075} & 0.8096 & 0.9766 & 0.9634 & 0.7686 & 0.6270 \\
\hline
\end{tabular}%
\end{table*}

\begin{table*}[t]
\centering
\caption{Comparison of Lightweight Retinal Vessel Segmentation Methods on FIVES Dataset}
\label{tab:fives_comparison}
\renewcommand{\arraystretch}{1.3}
\begin{tabular}{|l|c|c|c|c|c|c|}
\hline
\textbf{Work} & \textbf{Params(M)} & \textbf{SE} & \textbf{SP} & \textbf{ACC} & \textbf{F1 / Dice} & \textbf{Jaccard / IoU} \\
\hline
UNet (2015)\cite{bib13}                     & 31.03 & \textbf{0.9199} & 0.9820 & 0.9773 & 0.8325 & 0.7354 \\
XceptionLFOR (2024)\cite{bib42}                          & --    & 0.9153 & --     & --     & \textbf{0.8923} & --     \\
Attention U-Net (ResNet101V2) (2024) \cite{bib43}          & --    & --     & --     & 0.8507 & 0.8385 & 0.7221 \\
Attention U-Net (ResNet152V2) (2024)\cite{bib43}          & --    & --     & --     & 0.8331 & 0.8369 & 0.7199 \\
Attention U-Net (ResNet50V2) (2024) \cite{bib43}           & --    & --     & --     & 0.8512 & 0.8353 & 0.7175 \\
\hline
\textbf{LightVesselNet (Ours)}        & \textbf{0.075} & 0.8634 & \textbf{0.9912} & \textbf{0.9830} & 0.8649 & \textbf{0.7693} \\
\hline
\end{tabular}%
\end{table*}

Scores reported for competing methods (except UNet) were taken directly from their respective publications, as none of the methods included in the comparison provides publicly available training code or model weights. A hyphen (-) indicates that the corresponding metric was not reported in the referenced manuscript.

On DRIVE, LightVesselNet attains competitive results compared to models with much higher parameter count. On STARE, due to low contrast and limited training data, absolute F1 and Jaccard scores are lower than those of LFRA-Net and LFA-Net. Nevertheless, LightVesselNet maintains competitive accuracy (0.9696) with the fewest parameters.

On CHASE\_DB1, LightVesselNet achieves the best sensitivity (0.8640) and accuracy (0.9738) among all compared methods, demonstrating strong detection of thin and low-contrast vessels. Although LFRA-Net yields higher F1 (0.8550 vs.\ 0.8181) and Jaccard scores, it requires $2.3\times$ more parameters.

On HRF, LightVesselNet achieves an accuracy of 0.9634 and a sensitivity of 0.8096, competitive with much larger models such as UNet (31.03 M) and GraphSeg (19.32 M). LHU-VT achieves the highest absolute scores on this dataset, but does so with 19.40 M parameters ($>$250$\times$ more than LightVesselNet). 

On FIVES, LightVesselNet delivers its strongest absolute performance across all datasets. It achieves the best specificity (0.9912), accuracy (0.9830), and Jaccard score (0.7693). The only method with a higher F1 score is XceptionLFOR (0.8923), but its parameter count is not reported, making a fair efficiency comparison impossible. 

Overall, LightVesselNet consistently delivers strong performance across all five benchmarks despite its extremely compact design. While it does not always achieve the highest absolute scores, it offers a robust balance between accuracy and model size that makes it well-suited for real-world deployment in resource-constrained settings.

\subsubsection{Pareto Frontier Analysis}

Pareto frontier analysis is used to evaluate the trade-off between model performance and efficiency. In medical image segmentation, it helps compare methods in terms of accuracy (e.g., sensitivity or F1-score) and model complexity (e.g., number of parameters). A method is considered Pareto-optimal if no other method achieves higher performance with fewer or equal parameters. This analysis provides a clear view of how well a model balances accuracy and efficiency, which is important for lightweight models intended for practical use.

\begin{figure*}[htbp]
    \centering
    \includegraphics[width=0.7\textwidth]{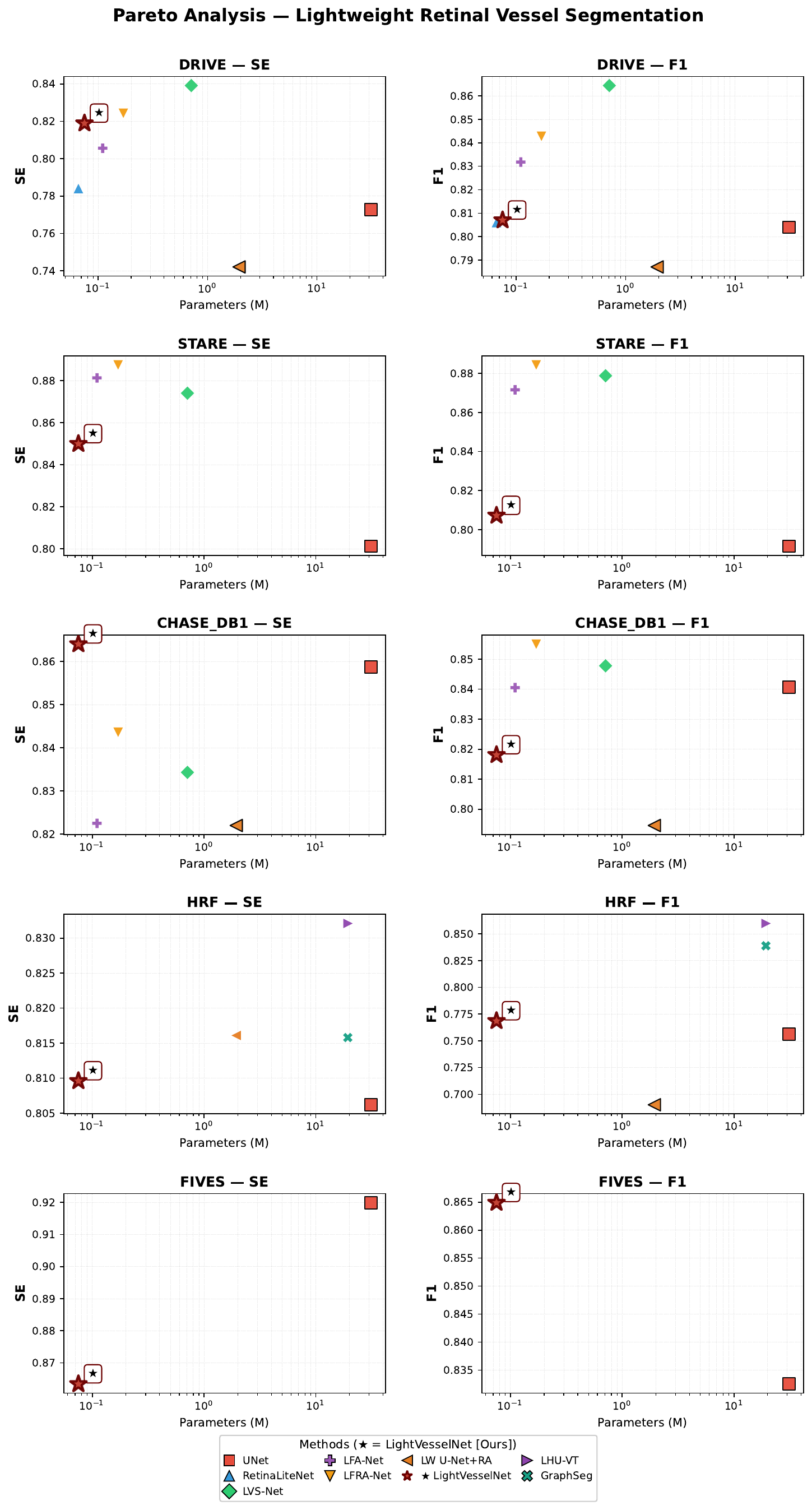}
    \caption{Pareto frontier analysis of lightweight retinal vessel segmentation methods across five datasets (DRIVE, STARE, CHASE\_DB1, HRF, FIVES). 
    Each row corresponds to one dataset with two columns: (1) Sensitivity (SE) vs. model parameters, and (2) F1-score vs. model parameters. The x-axis is plotted on a logarithmic scale. 
    LightVesselNet achieves competitive accuracy with significantly fewer parameters, positioning it on or near the Pareto frontier in most cases.}
    \label{fig:pareto_analysis}
\end{figure*}

Figure \ref{fig:pareto_analysis} illustrates the Pareto frontier analysis of lightweight retinal vessel segmentation methods across five benchmark datasets (DRIVE, STARE, CHASE\_DB1, HRF, and FIVES).

LightVesselNet consistently occupies the Pareto-optimal region across all five benchmarks. While it does not always achieve the highest absolute scores, it offers a robust balance between accuracy and efficiency that makes it well-suited for real-world deployment in resource-constrained settings.

A direct statistical comparison with competing methods is not possible in this study, as none of the methods included in the comparison provides publicly available training code, model weights, per-fold scores, or prediction masks.

\section{Conclusion}
\label{sec:Conclusion}

In this paper, we introduced LightVesselNet, an ultra-lightweight, sub-100K parameter neural network optimised for retinal blood vessel segmentation in resource-constrained environments. By proposing specialised components such as the MicroBlockSE for efficient depthwise separable convolutions and channel attention, the Multi-Scale Feature Aggregation (MSFA) bottleneck module for capturing variable vessel calibres, and an edge residual pathway to preserve low-level boundaries, the network minimises computational overhead without sacrificing clinical relevance. Extensive empirical evaluations across multiple benchmark datasets, such as DRIVE, STARE, CHASE\_DB1, FIVES, and HRF, confirm that LightVesselNet delivers competitive segmentation performance. Comprising only 75K parameters, it occupied the Pareto-optimal region in its parameter range. Furthermore, cross-dataset experiments validated its strong generalisation capability on unseen fundus scans, highlighting its resilience to domain shifts. Future work will explore further compression of the network via structured pruning and low-bit quantisation to reduce inference latency on low-power embedded architectures. Additionally, the applicability of the proposed architecture to broader medical image segmentation tasks will be investigated to evaluate its generalisability beyond retinal vessel segmentation. Future possible work might include statistical significance testing if there are works with publicly available weights or code. In summary, LightVesselNet achieves a favourable accuracy–efficiency trade-off, positioning it as a viable solution for deployment on mobile screening tools and edge diagnostic devices in remote or resource-constrained clinical environments.

\section*{Code Availability}
All the datasets used in this work are publicly available. The source code of the proposed model is available online for reproducibility at: \href{https://github.com/ShadmanSobhan/LightVesselNet}{LightVesselNet Repository}.

\end{document}